\documentclass[conference]{IEEEtran}
\IEEEoverridecommandlockouts

\usepackage{cite}
\usepackage{amsmath,amssymb,amsfonts}
\usepackage{algorithmic}
\usepackage{tikz}
\usepackage{graphicx}
\usepackage{booktabs}
\usepackage{textcomp}
\usepackage{xcolor}
\def\BibTeX{{\rm B\kern-.05em{\sc i\kern-.025em b}\kern-.08em
    T\kern-.1667em\lower.7ex\hbox{E}\kern-.125emX}}
\begin{document}

\title{Intuitionistic Fuzzy Graph Embedded Random Vector Functional Link with Multiview Learning}

\author{
\IEEEauthorblockN{Vrushank Ahire \hspace{2.5cm} Yogesh Kumar \hspace{2.5cm} M.A. Ganaie}
\IEEEauthorblockA{\textit{2022csb1002@iitrpr.ac.in} \hspace{1cm} \textit{yogesh.23csz0014@iitrpr.ac.in} \hspace{1cm} \textit{mudasir@iitrpr.ac.in}}
\vspace{0.5cm} 

\IEEEauthorblockA{
Dept. of Computer Science and Engineering, IIT Ropar, Punjab, India}
}

\maketitle

\begin{abstract}
Random Vector Functional Link (RVFL) networks are popular due to their fast training and universal approximation capabilities. However, RVFL models face challenges in preserving geometric relationships and utilizing multiple feature views effectively. To address these limitations we propose the Intuitionistic Fuzzy Graph Embedded Random Vector Functional Link with Multiview Learning (IFGRVFL-MV) model. The proposed approach comprises three key components: intuitionistic fuzzy sets for uncertainty handling, graph embedding to capture intrinsic geometric structures, and multiview learning to use complementary information from multiple feature spaces. The model assigns intuitionistic fuzzy membership and non-membership values to data points making it robust to outliers. Also, the graph embedding framework preserves topological structures, increasing the generalization performance. We performed experiments on benchmark datasets from UCI and KEEL repositories which concludes that IFGRVFL-MV outperforms existing models in classification accuracy. Our results establish that IFGRVFL-MV is a promising advancement in the domain of uncertainty and multiview environments. \end{abstract}


\section{Introduction}

Machine learning has evolved significantly over the years, with traditional algorithms such as Support Vector Machines (SVMs) \cite{hearst1998support}, Decision Trees, and k-Nearest Neighbors (k-NN) being widely used for classification and regression tasks. However, these methods often face challenges such as high computational complexity, sensitivity to hyperparameters, and difficulties in handling noisy or uncertain data \cite{hajiloo2013fuzzy}. For instance, SVMs, while effective for binary classification, struggle with large-scale datasets due to their quadratic optimization complexity \cite{khemchandani2007twin}. Similarly, traditional neural networks trained using backpropagation are prone to issues like slow convergence, local minima, and overfitting \cite{malik2023random}.

In contrast, Randomized Neural Networks (RNNs), particularly the RVFL network, have emerged as a promising alternative due to their simplicity, fast training speed, and universal approximation capabilities \cite{pao1994learning}. Unlike traditional neural networks, RVFL networks randomly initialize the weights of the hidden layer and compute the output weights analytically, avoiding the need for iterative optimization \cite{zhang2020new}. This approach not only reduces training time but also eliminates the risk of getting stuck in local minima. Furthermore, the direct links between the input and output layers in RVFL networks enhance their generalization performance, making them suitable for a wide range of applications, including healthcare, image classification and forecasting \cite{ren2016random}.

Recent research has focused on enhancing the capabilities of RVFL networks by incorporating advanced techniques such as fuzzy logic, Graph Embedding (GE), and Multiview Learning (MVL). For instance, the Intuitionistic Fuzzy RVFL (IFRVFL) model integrates intuitionistic fuzzy sets to handle uncertainty and imprecision in data, assigning membership and non-membership values to each sample based on its distance from the class center and the heterogeneity of its neighbors \cite{mishra2023intuitionistic}. This approach improves the model's robustness to noise and outliers, making it particularly effective for noisy datasets. Similarly, the Graph Embedded RVFL (GE-RVFL) model utilizes the GE framework to capture the geometric relationships within the data, preserving the topological structure and enhancing generalization performance \cite{yan2006graph}.

The concept of MVL has also been integrated with RVFL networks to utilize complementary information from multiple feature sets. MVL models, such as the Multiview RVFL (MvRVFL), combine data from different views to improve classification accuracy and robustness. For example, in image classification, features like color, texture, and shape can provide distinct yet complementary information, which MVL models effectively exploit. The Graph Random Vector Functional Link based on Multi-View Learning (GRVFL-MV) \cite{quadir2024multiview} is a notable example that combines RVFL, MVL, and the GE framework to achieve superior performance on diverse datasets.

Despite these advancements, challenges remain in handling noisy environments and effectively combining multiple techniques. Intuitionistic fuzzy sets have been shown to address uncertainty and noise effectively by assigning appropriate weights to samples based on their membership and non-membership values \cite{rezvani2019intuitionistic}. For instance, the Intuitionistic Fuzzy Twin Support Vector Machine (IFTSVM) combines intuitionistic fuzzy theory with twin SVMs to reduce the impact of noise and improve classification accuracy \cite{rezvani2019intuitionistic}. Similarly, the Graph Embedded Intuitionistic Fuzzy RVFL (GE-IFRVFL) model integrates GE and intuitionistic fuzzy sets to preserve the geometric structure of the data while handling uncertainty \cite{ganaie2024graph}.

However, existing models often focus on either fuzzy logic, GE, or MVL in isolation, leaving room for improvement in combining these techniques effectively. For example, while GE-IFRVFL excels in handling uncertainty and preserving geometric relationships, it does not fully exploit the potential of MVL. Similarly, GRVFL-MV utilizes MVL and GE but does not incorporate intuitionistic fuzzy sets to handle uncertainty and noise. This gap motivates the development of a unified model that integrates intuitionistic fuzzy theory, GE, and MVL into the RVFL framework.

In this paper, we propose the Intuitionistic Fuzzy Graph Embedded Random Vector Functional Link with Multiview (IFGRVFL-MV) model, which addresses the limitations of existing approaches by combining intuitionistic fuzzy sets, GE, and MVL. The proposed IFGRVFL-MV model assigns intuitionistic fuzzy membership and non-membership values to each data point, enabling it to handle noisy and uncertain datasets effectively. Additionally, the IFGRVFL-MV model utilizes the GE framework to capture the geometric relationships within the data, further enhancing its generalization performance. By integrating MVL, the model can exploit complementary information from multiple feature sets, leading to superior classification accuracy.

The key contributions of this paper are as follows:
\begin{itemize}
    \item We introduce intuitionistic fuzzy sets into the RVFL framework, enabling the model to handle uncertainty and imprecision in the data effectively. The membership and non-membership values are assigned based on the distance of each sample from the class center and the heterogeneity of its neighboring points, respectively.
    \item We incorporate the GE framework into the IFGRVFL-MV model to capture the geometric relationships within the data. By defining intrinsic and penalty graphs over the concatenated feature matrix, the model preserves the topological structure of the data, leading to better generalization performance.
    \item We utilizes MVL to combine information from multiple views of the data. This approach allows the model to capture complex patterns and relationships that may not be apparent in a single view, further enhancing its classification performance.
    \item We conduct experiments on benchmark datasets, including UCI and KEEL to evaluate the performance of the IFGRVFL-MV model. We perform statistical analyses, including the Friedman, Wilcoxon and win-tie-loss tests, to validate the superiority of the proposed model over existing approaches. The results confirm that the IFGRVFL-MV model achieves statistically significant improvements in classification performance.
\end{itemize}

The rest of the paper is organized as follows: Section II provides a detailed review of related works, including the standard RVFL model, intuitionistic fuzzy sets, and GE techniques. Section III presents the proposed IFGRVFL-MV model, including its mathematical formulation and optimization. Section IV discusses the experimental setup and results, followed by a conclusion and future directions in Section V.
\section{Related Works}

\subsection{Notations}

In this subsection, we introduce the mathematical notations used throughout the paper. Let \( z_i \) represent the \( i \)-th data point, where \( i = 1, 2, \dots, n \), and \( n \) is the total number of samples. Each data point \( z_i \) is represented in two views: \( P \) and \( Q \).  The feature vector in view \( P \) is denoted as \( z_i^P \in \mathbb{R}^{1 \times m_1} \), where \( m_1 \) is the number of features per sample in view \( P \), and the input matrix for view \( P \) is \( Z^P \in \mathbb{R}^{n \times m_1} \). Similarly, the feature vector in view \( Q \) is \( z_i^Q \in \mathbb{R}^{1 \times m_2} \), where \( m_2 \) is the number of features per sample in view \( Q \), and the input matrix for view \( Q \) is \( Z^Q \in \mathbb{R}^{n \times m_2} \). The target variable for the \( i \)-th data point is \( y_i \in \{+1, -1\} \), and the target vector for all data points is $Y = [y_1, y_2, \dots, y_n]^\top \in \mathbb{R}^{n \times 1}$.

\subsection{Preliminaries}

\subsubsection{Random Vector Functional Link Network}

The RVFL \cite{pao1994learning} network is a variation of single-layer feedforward networks designed to efficiently handle regression and classification tasks. Unlike traditional neural networks, RVFL initializes the weights connecting the input to the hidden layer randomly and keeps them fixed during training, enhancing computational efficiency and capturing both linear and non-linear data relationships. Its architecture comprises an input layer, a hidden layer with randomly initialized neurons, and an output layer. The hidden layer applies a specified activation function to transform the inputs, introducing non-linearities while reducing overfitting risks through its random connectivity. Additionally, direct connections from the input layer to the output layer bypass backpropagation, ensuring faster convergence and facilitating robust information flow.
\begin{figure}[h!]
    \centering
    \begin{tikzpicture}
        \node[draw, line width=0.1mm, rectangle, inner sep=5pt] 
        at (0,0) {\includegraphics[width=0.25\textwidth]{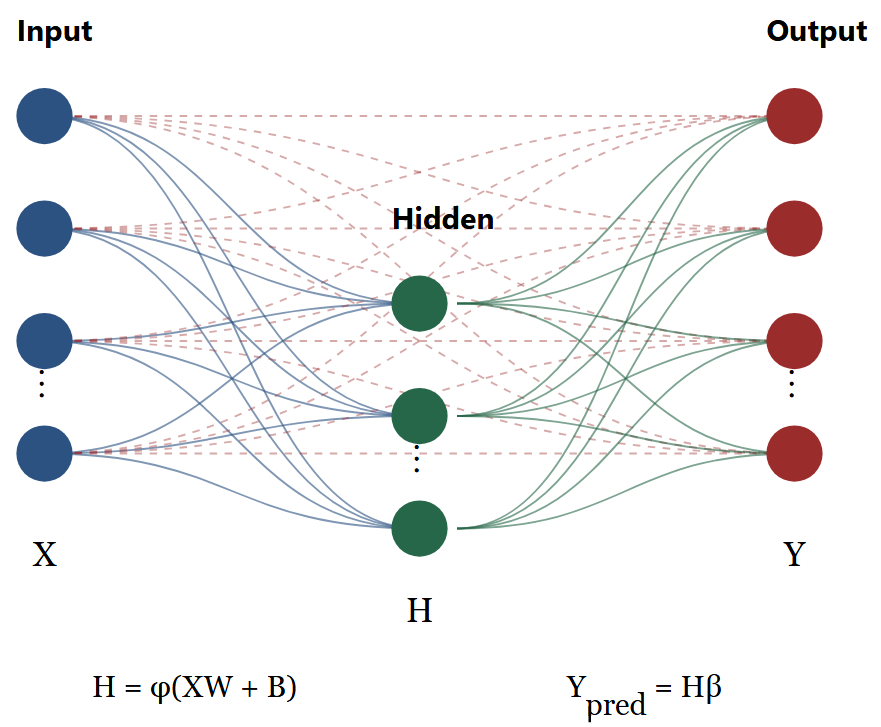}};
    \end{tikzpicture}
    \caption{RVFL Network Architecture}
    \label{fig:RVFL_Image}
\end{figure}
Let \( Z \in \mathbb{R}^{n \times m_1} \) be the input data matrix, where \( n \) is the number of samples and \( m_1 \) is the number of features. The hidden layer is constructed by randomly generating the weights \( W \in \mathbb{R}^{m_1 \times h} \) and biases \( B \in \mathbb{R}^{n \times h} \), where \( h \) is the number of hidden neurons. The weight matrix  \( W \) are typically drawn from a uniform distribution in the range \([-1, 1]\), and the bias matrix \( B \) is also randomly initialized.

The output of the hidden layer matrix, denoted as \( H \in \mathbb{R}^{n \times h} \), is computed by applying a nonlinear activation function \( \phi(\cdot) \) to the linear combination of the input features and the randomly generated weights and biases:
\begin{equation}
H = \phi(Z W + B),
\end{equation}
where \( \phi(\cdot) \) is leaky Relu activation function. The matrix \( H \) represents the transformed features in the hidden layer.

In RVFL, the original input features \( Z \) are directly passed to the output layer, along with the transformed features from the hidden layer \( H \). This is achieved by concatenating \( Z \) and \( H \) to form a new matrix \( H_{\text{c}} \in \mathbb{R}^{n \times (m_1 + h)} \):
\begin{equation}
H_{\text{concat}} = [Z \quad H].
\end{equation}

The output layer weights \( \beta \in \mathbb{R}^{(m_1 + h) \times 1} \) are then computed using a closed-form solution, typically via the least squares method or the Moore-Penrose pseudoinverse. The predicted output \( Y_{\text{pred}} \in \mathbb{R}^{n \times 1} \) is given by:
\begin{equation}
Y_{\text{pred}} = H_{\text{c}} \beta.
\end{equation}

The optimization problem for RVFL can be formulated as:
\begin{align}
&\min_{\beta} \frac{1}{2} \| \beta \|^2 + \frac{c}{2} \| \xi \|^2, \\
&\text{subject to:} \quad H_{\text{c}} \beta - Y = \xi \nonumber
\end{align}
where \( Y \in \mathbb{R}^{n \times 1} \) is the true target vector, \( \xi \in \mathbb{R}^{n \times 1} \) is the error term, and \( c \) is a regularization parameter that controls the trade-off between the norm of the weights and the error.

The solution to this optimization problem is given by:
\begin{equation}
\beta = 
\begin{cases}
(H_{\text{c}}^\top H_{\text{c}} + \frac{1}{c} I)^{-1} H_{\text{c}}^\top Y, & \text{if } (m + h) \leq n, \\[10pt]
H_{\text{c}}^\top (H_{\text{c}} H_{\text{c}}^\top + \frac{1}{c} I)^{-1} Y, & \text{if } n < (m + h),
\end{cases}
\end{equation}
where \( I \) is the identity matrix of appropriate dimensions.\\

\subsubsection{Multiview Learning}

MVL \cite{zhao2017multi} is a machine learning paradigm that enhances model performance by leveraging multiple complementary representations, or ``views," of the same data. Unlike traditional approaches that concatenate features into a single representation, MVL optimizes separate functions for each view while jointly integrating them to improve generalization. This approach addresses challenges such as high-dimensional sparsity and the loss of unique statistical properties inherent in single-view methods.

\begin{itemize}
    \item Co-training: Ensures agreement between views to reinforce predictions and improve confidence, as introduced by the Co-training model \cite{blum1998combining}.
    \item Co-regularization: Aligns learning processes to reduce discrepancies between views, such as in graph-based and kernel-based models \cite{sindhwani2005co}, \cite{sun2013online}.
    \item Margin consistency: Preserves stable decision boundaries across views, exemplified by the SVM-2K model \cite{farquhar2005two}.
\end{itemize}

The effectiveness of MVL has been demonstrated in diverse applications. In image classification, multi-view semi-supervised learning framework that iteratively trains view-specific classifiers with labeled and pseudo-labeled images for effective image classification \cite{zhu2016multi}. In biomedical research, it integrates clinical and imaging data for tasks such as predicting disease progression \cite{zhao2022multi}. Similarly, MVL has been applied in financial distress prediction \cite{duarte2020forecasting} and consumer behavior analysis, where the integration of heterogeneous data provides deeper insights \cite{zhang2022detecting}.

Mathematically, MVL seeks to jointly optimize models for multiple views while preserving their complementary nature. Let \( Z^P \in \mathbb{R}^{n \times m_1} \) and \( Z^Q \in \mathbb{R}^{n \times m_2} \) represent two distinct views of the data, where \( n \) is the number of samples, \( m_1 \) and \( m_2 \) are the dimensions of the features in views \( P \) and \( Q \), respectively. The objective of MVL can often be expressed as:
\begin{equation}
\min_{f_P, f_Q} \mathcal{L}(f_P(Z^P), f_Q(Z^Q)) + \lambda \mathcal{R}(f_P, f_Q),
\end{equation}
where \( f_P \) and \( f_Q \) are view-specific functions, \( \mathcal{L}(\cdot) \) is the loss function measuring the disagreement between the views, \( \mathcal{R}(\cdot) \) is a regularization term to align the functions, and \( \lambda > 0 \) controls the trade-off between agreement and individual function optimization.

One of the simplest yet powerful strategies for MVL is early fusion, where the views are combined at the feature level into a unified representation:
\begin{equation}
Z = [Z^P \quad Z^Q]
\end{equation}
In this approach, the fused representation $Z\in \mathbb{R}^{n \times (m_1 + m_2)}$ consolidates all available information into a single matrix, allowing the model to utilize the interactions between features from both views directly. The optimization objective for early fusion can then be reformulated as:
\begin{equation}
\min_f \mathcal{L}(f(Z), Y) + \lambda \mathcal{R}(f),
\end{equation}
where \( f \) is the prediction function applied to the unified representation \( Z \), and \( Y \) is the target variable.

While concerns such as the curse of dimensionality are often raised for early fusion strategies, these challenges can be effectively mitigated through dimensionality reduction techniques like principal component analysis \cite{mackiewicz1993principal} or sparse representation learning. Moreover, the fused representation enables the model to explore rich feature interactions across views, often leading to improved predictive performance when the views are highly complementary.

\subsubsection{Intuitionistic Fuzzy}

SVMs have demonstrated impressive performance in classification tasks; however, their sensitivity to noise and outliers presents significant challenges \cite{cristianini2000introduction}. To mitigate these issues, researchers have explored fuzzy-based approaches. Intuitionistic fuzzy sets, proposed by Atanassov \cite{atanassov1999intuitionistic}, provide an enhanced framework by incorporating both membership and non-membership degrees to represent uncertainties in data more comprehensively. This representation is particularly effective near decision boundaries, where ambiguity often arises. Studies, including those by Rezvani et al. \cite{rezvani2019intuitionistic}, show that integrating intuitionistic fuzzy logic into SVMs improves their robustness and accuracy in the presence of noisy data, leading to more reliable classification outcomes. Overall, the adoption of Intuitionistic Fuzzy Twin Support Vector Machines (IFT-SVM) exemplifies a promising advance in the quest for effective machine learning solutions, particularly in complex environments rife with uncertainty. 

The membership function \( f(z_i) \) assigns a weight to each sample based on its proximity to the class centers. Specifically, for a sample \( z_i \) with target \( y_i \), the membership function is defined as:

\begin{equation}
f(z_i) = 
\begin{cases} 
1 - \frac{\| \theta(z_i) - c^+ \|}{r^+ + c}, & \text{if } y_i = +1, \\
1 - \frac{\| \theta(z_i) - c^- \|}{r^- + c}, & \text{if } y_i = -1,
\end{cases}
\label{eq:membership}
\end{equation}

where \( c^+ \) and \( c^- \) are the class centers for the positive and negative classes, respectively, and \( r^+ \) and \( r^- \) are their respective radii. The parameter \( c \) is a small positive constant, and \( \theta(z_i) \) represents a nonlinear mapping of the data. The Frobenius norm is used to calculate the distance.

The class centers \( c^+ \) and \( c^- \) are calculated as the averages of the feature vectors for each class:

\begin{equation}
c^+ = \frac{1}{n_+} \sum_{y_i = +1} \theta(z_i), \quad c^- = \frac{1}{n_-} \sum_{y_i = -1} \theta(z_i),
\end{equation}

where \( n_+ \) and \( n_- \) are the numbers of positive and negative samples, respectively. The radii for each class are given by:

\begin{equation}
r^\pm = \max_{y_i = \pm 1} \| \theta(z_i) - c^\pm \|.
\end{equation}

The non-membership function \( f^*(z_i) \) captures the degree of non-membership of a sample. It is determined by the ratio of heterogeneous points (those with a different target class) to the total number of neighboring points:

\begin{equation}
f^*(z_i) = (1 - f(z_i)) \Phi(z_i),
\label{eq:nonmembership}
\end{equation}

where \( \Phi(z_i) \) is defined as:

\begin{equation}
\Phi(z_i) = \frac{|\{ z_j : \| \theta(z_i) - \theta(z_j) \| \leq \eta, y_j \neq y_i \}|}{|\{ z_j : \| \theta(z_i) - \theta(z_j) \| \leq \eta \}|}.
\label{eq:phi}
\end{equation}

Here, \( \eta \) is a parameter that controls the proximity threshold, and \( |\cdot| \) denotes the cardinality of a set.

To facilitate decision-making, a score function \( s_i \) is used, which is based on both the membership and non-membership values of each sample. The score function is defined as:

\begin{equation}
s_i =
\begin{cases} 
f_i, & \text{if } f^*_i = 0, \\
0, & \text{if } f_i \leq f^*_i, \\
\frac{1 - f^*_i}{k - Y_i - f^*_i}, & \text{otherwise}.
\end{cases}
\label{eq:score}
\end{equation}

In this formulation, \( s_i \) is used to quantify the relative importance of each sample in the classification task.

Finally, the distance between two samples, \( z \) and \( x \), can be computed using the kernel function \( \mathcal{K}(z, x) \). The distance is given by:

\begin{equation}
\| \theta(z) - \theta(x) \| = \sqrt{\mathcal{K}(x, x) + \mathcal{K}(z, z) - 2\mathcal{K}(z, x)}.
\label{eq:kerneldist}
\end{equation}

Additionally, the distance between a sample \( z_i \) and its class centers \( c^\pm \) can be calculated as:

\begin{align}
d^\pm &= \| \theta(z_i) - c^\pm \| \label{eq:classdist} \\
&= \sqrt{\mathcal{K}_{ii} + \frac{1}{n_\pm^2} \sum_{y_m = \pm 1} \sum_{y_n = \pm 1} \mathcal{K}_{nm} - \frac{2}{n_\pm} \sum_{y_j = \pm 1} \mathcal{K}_{ij}}.\nonumber
\end{align}

where \( \mathcal{K}_{ii} = \mathcal{K}(z_i, z_i) \) and \( n_\pm \) is the number of samples in the positive or negative class.

In summary, the intuitionistic fuzzy membership scheme provides a robust framework for classifying data by addressing uncertainty. By utilizing membership and non-membership functions as described in equations \eqref{eq:membership}, \eqref{eq:nonmembership}, and \eqref{eq:phi}, along with the score function \eqref{eq:score} and distance calculations \eqref{eq:kerneldist}--\eqref{eq:classdist}, the model improves the accuracy and stability of classification tasks in uncertain environments.

\subsection{Graph Embedding Framework}

GE \cite{yan2006graph} is a framework designed to represent data's graphical structure in a vector space using intrinsic and penalty graphs. The intrinsic graph \(\mathcal{G}^{int} = \{Z, \Omega^{int}\}\) captures relationships between data samples, while the penalty graph \(\mathcal{G}^{pen} = \{Z, \Omega^{pen}\}\) imposes penalties on specific relationships. Here, \(Z = \{z_1, z_2, \dots, z_n\}\) represents the data samples.

The framework uses two weight matrices:
\begin{itemize}
    \item Similarity weight matrix \(\Omega^{int} \in \mathbb{R}^{n \times n}\): Encodes pairwise relationships between vertices in \(X\).
    \item Penalty weight matrix \(\Omega^{pen} \in \mathbb{R}^{n \times n}\): Assigns penalties to specific vertex relationships.
\end{itemize}

The optimization problem in GE is formulated as:
\begin{equation}
w^* = \mathop{\arg\min}_{\text{tr}(w_0^T Z^T P Z w_0) = q} \sum_{i \neq j} \left\| w_0^T z_i - w_0^T z_j \right\|^2 \Omega^{int}_{ij} 
\end{equation}
This can be rewritten using the trace operator:
\begin{equation}
w^* = \mathop{\arg\min}_{\text{tr}(w_0^T Z^T P Z w_0) = q} \text{tr}(w_0^T Z^T M Z w_0)
\end{equation}
where, \(M = D - \Omega^{int}\) is the graph matrix of the intrinsic graph, where \(D\) is a diagonal matrix with \(D_{ii} = \sum_j\Omega^{int}_{ij}\).\\
\(P\) is either a diagonal normalization matrix or the graph matrix of the penalty graph, \(P = M^p = D^p - \Omega^{pen}\), and \(q\) is a constant.

The solution is obtained by solving the generalized eigenvalue problem:
\begin{equation}
    A_i z = \lambda A_p z
\end{equation}
where: \( A_i = Z^T M Z \), \( A_p = Z^T P Z \)

The transformation matrix is formed from the eigenvectors of:
\begin{equation}
    A_w = A_p^{-1} A_i,
\end{equation}
which integrates relationships from both the intrinsic and penalty graphs. This framework effectively reduces dimensionality while preserving the structural relationships and penalties in the data.

\subsection{Graph Embedded Intuitionistic Fuzzy Weighted RVFL Network:}
The Graph Embedded Intuitionistic Fuzzy Weighted RVFL (GE-IFWRVFL) \cite{malik2022graph} network integrates intuitionistic fuzzy sets and GE into the RVFL framework \cite{malik2022graph}. The model assigns intuitionistic fuzzy weights to data points based on their membership and non-membership values, and it preserves the geometric structure of the data using GE. The optimization problem for GE-IFWRVFL is formulated as:

\begin{align}
& \min_{\beta, \xi} \quad \frac{1}{2} \lambda \| S^{\frac{1}{2}} \xi \|_2^2 + \frac{1}{2} \| \beta \|_2^2 + \frac{1}{2} \alpha \| A^{\frac{1}{2}}_w \beta \|_2^2 \\
& \text{subject to: } \quad H_{c}\beta - Y = \xi \nonumber
\end{align}
where, $\lambda$ and $\alpha$ are tunable parameters, $S$ is a diagonal matrix of score values, $\xi$ is the error matrix, $\beta$ represents the output layer weights and $A_w$ is the GE matrix.

The Lagrange's function for the optimization problem is:
\begin{equation}
    L = \frac{1}{2} \lambda \| S^{\frac{1}{2}} (H_c\beta - Y) \|_2^2 + \frac{1}{2} \| \beta \|_2^2 + \frac{1}{2} \alpha \| A^{\frac{1}{2}}_w \beta \|_2^2.
\end{equation}

The output parameter $\beta$ is derived as:
\begin{equation}
    \beta = \left( H_c^T S H_c + \frac{1}{\lambda} I + \frac{\alpha}{\lambda} A_w \right)^{-1} H_c^T S Y.
\end{equation}

For a test sample $z$, the final decision is given by:
\begin{equation}
    g(x) = [z, h(z)] \beta.
\end{equation}
This model effectively handles uncertainty and noise while preserving the geometric relationships within the data.

\subsection{Graph Random Vector Functional Link Neural Network based on Multi-View Learning:}

The Graph Random Vector Functional Link Neural Network based on Multi-View Learning (GRVFL-MV) \cite{tanveer2024grvfl} combines the RVFL network with Multi-View Learning (MVL) and GE. This model uses the GE framework to represent multi-view data while preserving its structure through Local Fisher Discriminant Analysis (LFDA). The optimization problem for the GRVFL-MV model is defined as follows:
\begin{align}
&\min_{\beta_{1}, \beta_{2}}  \quad \frac{c_{1}}{2} \|\xi_{1}\|^{2}_{2} + \frac{c_{2}}{2} \|\xi_{2}\|^{2}_{2} + \frac{c_{3}}{2} \|\beta_{1}\|^{2}_{2} + \frac{1}{2} \|\beta_{2}\|^{2}_{2} \nonumber \\
& \quad \quad \quad+ \frac{\theta_{1}}{2} \|A^{1/2}_{w_1} \beta_{1}\|^{2}_{2} + \frac{\theta_{2}}{2} \|A^{1/2}_{w_2} \beta_{2}\|^{2}_{2} + \rho \xi^{\dagger}_{1} \xi_{2} \nonumber \\
& \text{subject to:}  \quad H_{c}^P\beta_{1} - Y = \xi_{1} \quad H_{c}^Q\beta_{2} - Y = \xi_{2}.
\end{align}
where, \(\beta_1, \beta_2\) are output weights; \(\xi_1, \xi_2\) are error terms; \(c_1, c_2, c_3, \theta_1, \theta_2, \rho\) are regularization parameters; \(A_{w_1}, A_{w_2}\) are GE weights; \(H_c^P = [Z^P \quad H^P]\) and \(H_c^Q = [Z^Q \quad H^Q]\), where \(H^P = \phi(Z^P W^P + B^P)\) and \(H^Q = \phi(Z^Q W^Q + B^Q)\), with \(\phi(\cdot)\) as the activation function. Here, \(Z^P, Z^Q\) are the input data for the two views, and \(W^P, B^P\), \(W^Q, B^Q\) are their respective randomly initialized weights and biases.

The Lagrangian is derived, and partial differentiation provides the weight matrices \(\beta_{1}\) and \(\beta_{2}\) by solving the matrix equation.

Assuming:
\[
\begin{aligned}
E_{11} &= c_{3}I_1 + \theta_{1}A_{w_1} + c_{1}({H_{c}^{P}})^TH_{c}^P, \\
E_{12} &=  \rho ({H_{c}^{P}})^TH_{c}^Q, \\
E_{21} &= \rho ({H_{c}^{Q}})^TH_{c}^P, \\
E_{22} &=  I_2 + \theta_{2}A_{w_2} + c_{2}({H_{c}^{Q}})^TH_{c}^Q.
\end{aligned}
\]

\begin{equation}
\begin{bmatrix}
\beta_{1} \\
\beta_{2}
\end{bmatrix}
=
\begin{bmatrix}
E_{11}
& E_{12} \\
E_{21}
& E_{22}
\end{bmatrix}^{-1}
\begin{bmatrix}
{({H_{c}}^P})^{T}(c_{1} + \rho) \\
{({H_{c}}^Q})^{T}(c_{2} + \rho)
\end{bmatrix}
Y
\end{equation}

For a new data point \(z\) with representations \(z^{P}\) and \(z^{Q}\) for view-P and view-Q, respectively, the classification function is:
\begin{equation}
    \text{class}(z) = \operatorname*{arg\,max}_{c_i \in \{+,-\}} \{y_{c_{i}}\}
\end{equation}
where the term for \( y_c \) is given by \\\( y_{c} = \frac{1}{2} \left( [z^P \;\; \phi(z^P W^P + b^P)] \beta_{1} + [z^Q \;\; \phi(z^Q W^Q + b^Q)] \beta_{2} \right) \).
\section{Proposed Model}
In this section, we present the Intuitionistic Fuzzy Graph Random Vector Functional Link with Multiview (IFGRVFL-MV) model. This model extends the Graph Random Vector Functional Link based on Multi-View Learning (GRVFL-MV) by incorporating Intuitionistic Fuzzy Sets (IFS). The proposed model utilizes the strengths of intuitionistic fuzzy theory, GE, and MVL to handle uncertainty, noise, and geometric relationships in the data.

The objective function becomes:
\begin{equation}
\begin{aligned}
&\min_{\beta_1, \beta_2} \quad  \frac{c_1}{2}||S_1^{1/2} \xi_1||^2_2 + \frac{c_2}{2}||S_2^{1/2} \xi_2||^2_2 + \frac{c_3}{2}||\beta_1||^2_2 + \frac{c_4}{2}||\beta_2||^2_2 \\
& \quad +\frac{\theta_1}{2}||A^{1/2}_{w_1} \beta_1||^2_2 + \frac{\theta_2}{2}||A^{1/2}_{w_2} \beta_2||^2_2 + \rho \xi_1^T S_1^{1/2} S_2^{1/2} \xi_2 \\
&\text{subject to} \quad H_c^P \beta_1 - Y = \xi_1 \quad \text{and} \quad H_c^Q \beta_2 - Y = \xi_2
\end{aligned}
\end{equation}
where the diagonal matrices $S_1$ and $S_2$ contain intuitionistic fuzzy weights $s_i^P$ and $s_i^Q$ for View-P and View-Q, and their square roots $S_1^{1/2}$, $S_2^{1/2}$ properly apply these weights in regularization, the terms 
$\frac{c_1}{2} \|S^{1/2}_1 \xi_1\|_2^2$ and $\frac{c_2}{2} \|S^{1/2}_2 \xi_2\|_2^2$ penalize the fuzzy-weighted errors for View-P and View-Q, 
$\frac{c_3}{2} \|\beta_1\|_2^2$ and $\frac{c_4}{2} \|\beta_2\|_2^2$ regularize weights to prevent overfitting, 
$\frac{\theta_1}{2} \|A^{1/2}_{w1} \beta_1\|_2^2$ and $\frac{\theta_2}{2} \|A^{1/2}_{w2} \beta_2\|_2^2$ preserve geometric structure, 
$\rho \xi_1^T S^{1/2}_1 S^{1/2}_2 \xi_2$ enforces cross-view error consistency, and the constraints 
$H_c^P \beta_1 - Y = \xi_1$, $H_c^Q \beta_2 - Y = \xi_2$ ensure predictions align with true labels.

The Lagrangian of the optimization problem for (IFGRVFL-MV) is:

\begin{equation}
    \begin{aligned}
    L = \; & \frac{c_1}{2}||S_1^{1/2} \xi_1||^2 + \frac{c_2}{2}||S_2^{1/2} \xi_2||^2 + \frac{c_3}{2}||\beta_1||^2 + \frac{c_4}{2}||\beta_2||^2 \\
    & + \frac{\theta_1}{2}||A^{1/2}_{w_1} \beta_1||^2_2 + \frac{\theta_2}{2}||A^{1/2}_{w_2} \beta_2||^2_2 + \rho \xi_1^T S_1^{1/2} S_2^{1/2} \xi_2 \\
    & - \alpha_1^T (H_c^P \beta_1 - Y - \xi_1) - \alpha_2^T (H_c^Q \beta_2 - Y - \xi_2),
    \end{aligned}
\end{equation}
where \(\alpha_1\) and \(\alpha_2\) are Lagrange multipliers of the appropriate dimensions.

The partial derivatives of the Lagrangian with respect to \(\xi_1, \xi_2, \beta_1, \beta_2, \alpha_1,\) and \(\alpha_2\) are computed as follows:

\begin{align}
    \frac{\partial L}{\partial \xi_1} &= c_1 S_1 \xi_1 + \rho S_1^{1/2} S_2^{1/2} \xi_2 + \alpha_1 = 0, \\
    \frac{\partial L}{\partial \xi_2} &= c_2 S_2 \xi_2 + \rho S_2^{1/2} S_1^{1/2} \xi_1 + \alpha_2 = 0, \\
    \frac{\partial L}{\partial \beta_1} &= c_3 \beta_1 + \theta_1 A_w^P \beta_1 - (H_c^P)^T \alpha_1 = 0, \\
    \frac{\partial L}{\partial \beta_2} &= c_4 \beta_2 + \theta_2 A_w^Q \beta_2 - (H_c^Q)^T \alpha_2 = 0, \\
    \frac{\partial L}{\partial \alpha_1} &= H_c^P \beta_1 - Y - \xi_1 = 0, \\
    \frac{\partial L}{\partial \alpha_2} &= H_c^Q \beta_2 - Y - \xi_2 = 0.
\end{align}

By substituting the partial derivatives, we obtain the following system of equations:

\begin{equation}
\resizebox{0.36\textwidth}{!}{$
\begin{aligned}
    &c_3 \beta_1 + \theta_1 A_w^P \beta_1 + (H_c^P)^T \big( c_1 S_1 (H_c^P \beta_1 - Y) \\
    &+ \rho S_1^{1/2} S_2^{1/2} (H_c^Q \beta_2 - Y) \big) = 0
\end{aligned}
$}
\end{equation}

\begin{equation}
\resizebox{0.36\textwidth}{!}{$
\begin{aligned}
    &c_4 \beta_2 + \theta_2 A_w^Q \beta_2 + (H_c^Q)^T  \big( c_2 S_2 (H_c^Q \beta_2 - Y) \\
    &+ \rho S_2^{1/2} S_1^{1/2} (H_c^P \beta_1 - Y) \big) = 0
\end{aligned}
$}
\end{equation}

Simplifying these equations, we get:
\begin{equation}
\resizebox{0.49\textwidth}{!}{$
\begin{aligned}
    &\left( c_3 I_1 + \theta_1 A_w^P + c_1 (H_c^P)^T S_1 H_c^P \right) \beta_1 + \left( \rho (H_c^P)^T S_1^{1/2} S_2^{1/2} H_c^Q \right) \beta_2 \\
    &= (H_c^P)^T \left( c_1 S_1 + \rho S_1^{1/2} S_2^{1/2} \right) Y,
\end{aligned}
$}
\end{equation}

\begin{equation}
\resizebox{0.49\textwidth}{!}{$
\begin{aligned}
    &\left( \rho (H_c^Q)^T S_2^{1/2} S_1^{1/2} H_c^P \right) \beta_1 + \left( c_4 I_2 + \theta_2 A_w^Q + c_2 (H_c^Q)^T S_2 H_c^Q \right) \beta_2 \\
    &= (H_c^Q)^T \left( c_2 S_2 + \rho S_2^{1/2} S_1^{1/2} \right) Y
\end{aligned}
$}
\end{equation}

where \(I_1\) and \(I_2\) are identity matrices of appropriate dimensions based on the matrices $A_w^P$ and $A_w^Q$ respectively.

The system of equations can be written in matrix form as:
Assuming:
\[
\begin{aligned}
E_{11} &= c_3 I_1 + \theta_1 A_w^P + c_1 (H_c^P)^T S_1 H_c^P, \\
E_{12} &=  \rho (H_c^P)^T S_1^{1/2} S_2^{1/2} H_c^Q, \\
E_{21} &= \rho (H_c^Q)^T S_2^{1/2} S_1^{1/2} H_c^P, \\
E_{22} &= c_4 I_2 + \theta_2 A_w^Q + c_2 (H_c^Q)^T S_2 H_c^Q.
\end{aligned}
\]

\begin{equation}
\begin{bmatrix}
\beta_1 \\
\beta_2
\end{bmatrix}
=
\begin{bmatrix}
E_{11}
& E_{12} \\
E_{21}
& E_{22}
\end{bmatrix}^{-1}
\begin{bmatrix}
(H_c^P)^T (c_1 S_1 + \rho S_1^{1/2} S_2^{1/2}) \\
(H_c^Q)^T (c_2 S_2 + \rho S_2^{1/2} S_1^{1/2})
\end{bmatrix}
Y
\end{equation}

The Intuitionistic Fuzzy Graph Random Vector Functional Link with Multiview (IFGRVFL-MV) model extends the original GRVFL-MV by incorporating intuitionistic fuzzy sets to handle uncertainty and noise in the data. The key highlights are as follows:
\begin{itemize}
    \item Intuitionistic fuzzy weighting of data points for each view
    \item Weighted optimization problem to reduce the impact of uncertain samples in each view
    \item Modified GE framework with intuitionistic fuzzy weights for each view
\end{itemize}

This formulation provides a robust framework for multi-view learning in uncertain environments, using the strengths of intuitionistic fuzzy sets to improve generalization and classification performance.

\section{Experiments and Results}
\subsection{Experimental Setup}
In this study, we introduce a modified GE framework that incorporates intuitionistic fuzzy weights to improve multi-view learning, especially in uncertain environments. In our experiment, we utilize an Intel Core i7-6700 CPU at 3.40GHz x8, running Ubuntu 22.04.3 LTS with Python 3.9. The framework employs a Gaussian kernel to compute similarity weights, defined as:

\begin{equation}
   \mathcal{K}(z,y) = \exp\left(-\frac{\|z - x\|^2}{\mu^2}\right)
\end{equation}
where \( \mu \) is a crucial kernel parameter that influences model performance. The datasets used in our study are sourced from the UCI \cite{asuncion2007uci} and KEEL \cite{alcal2011keel} repositories, which include a diverse range of classification tasks. These datasets are split into 70\% for training and 30\% for testing to ensure a fair evaluation. To optimize hyperparameters, we use a grid search method with five-fold cross-validation, where each dataset is split into five subsets, training on four and testing on one in an iterative manner. This technique ensures a robust and unbiased performance assessment.

\subsection{Time Complexity Analysis}
The computational complexity of the IFGRVFL-MV model extends that of GRVFL-MV by incorporating intuitionistic fuzzy computations. The GE step for \( Z^P \) and \( Z^Q \) incurs a complexity of \( O\left((m_1+h_1)^3 + (m_1+h_1)^2n\right) + O\left((m_2+h_2)^3 + (m_2+h_2)^2n\right) \). The intuitionistic fuzzy component introduces additional calculations for membership and non-membership functions, including class center and radius computation, leading to an extra \( O(n/2) \) complexity, which is relatively small and can be considered negligible. Solving the optimization problem requires inverting a square matrix of order \( (m_1+m_2+h_1+h_2) \), contributing a complexity of \( O\left((m_1+m_2+h_1+h_2)^3\right) \). Combining these, the total complexity of IFGRVFL-MV is \( O\left((m_1+m_2+h_1+h_2)^3 + \frac{n}{2}\right) \) with the dominant cost being determined by the GE and matrix inversion, while the fuzzy clustering computations contribute a negligible additional cost.

\begin{table*}[!htb]
\caption{{Comparison of AUC values for different models across various datasets, including optimized parameters and average performance metrics.}}
    \centering
    {\renewcommand{\arraystretch}{1.2}
    \resizebox{\textwidth}{!}{
    \begin{tabular}{|c|c|c|c|c|c|}
    \hline
    \textbf{Dataset}   & \textbf{RVFL1} \cite{pao1994learning} & \textbf{RVFL2} \cite{pao1994learning} & \textbf{GE-IFWRVFL} \cite{malik2022graph} & \textbf{GRVFL-MV}\cite{tanveer2024grvfl} & \textbf{IFGRVFL-MV}\\  
     & \textbf{ Acc $\uparrow$} & \textbf{ Acc $\uparrow$} & \textbf{ Acc $\uparrow$} & \textbf{ Acc $\uparrow$}& \textbf{ Acc $\uparrow$}\\ 

    \textbf{$(Patterns \times Features)$} & \textbf{($\mathbf{h, c}$)} & \textbf{($\mathbf{h, c}$)}& \textbf{($\mathbf{h, c, \alpha, \mu}$)} & \textbf{($\mathbf{h, c, \rho, \theta}$)} & \textbf{($\mathbf{h, c, \rho, \theta, \mu}$)} \\\hline

    pittsburg-bridges-T-OR-D&\textbf{87.09}&77.41&82.51&84.74&\textbf{87.09}\\($102 \times 8$)&(163, 1)&(83, 10)&(83, 10000, 100, 0.0001)&(123, 0.00001, 0.001, 0.00001)&(63, 0.001, 0.00001, 0.00001, 10)\\ \hline
    breast-cancer&82.55&82.55&87.56&\textbf{94.18}&93.34\\
    ($286\times10$)&(123,100)&(183,100)&(103,10,1,10000)&(143, 0.01, 0.001, 0.00001)&(23, 0.001, 0.00001, 0.00001, 0.00001)\\ \hline
    planning&72.72&70.90&71.26&72.72&\textbf{73.90}\\($182\times13$)&(3, 10)&(3, 0.00001)&(83, 10000, 100000, 100000)&(3, 0.00001, 0.1, 100)&(83, 0.00001, 0.00001, 0.00001, 1)\\ \hline
    mammographic&61.59&62.97&71.50&78.89&\textbf{83.39}\\($961 \times 6$)&(203, 10000)&(143, 1000)&(3, 1, 100, 1)&(43, 0.001, 0.0001, 0.0001)&(63, 0.0001, 0.00001, 0.0001, 1)\\ \hline
    breast-cancer-wisc-prog&81.66&75&79.02&78.33&\textbf{81.92}\\($198 \times 34$)&(183, 0.1)&(3, 0.1)&(63, 100000, 0.0001, 10000)&(3, 0.001, 0.0001, 0.00001)&(23, 0.00001, 0.1, 10, 0.1)\\ \hline
    Pima&67.09&67.96&70.17&73.59&\textbf{74.16}\\($768 \times 9$)&(163, 10)&(183, 100)&(3, 0.0001, 100, 1)&(43, 0.01, 0.0001, 0.00001)&(23, 0.00001, 0.0001, 0.01, 0.00001)\\ \hline
    cylinder-bands&54.54&55.19&64.49&69.48&\textbf{70.03}\\($512 \times 36$)&(183, 10)&(183, 1)&(203, 10000, 1000, 0.1)&(123, 0.1, 0.01, 0.01)&(3, 100000, 0.1, 1000, 0.1)\\ \hline
    checkerboard&68.59&71.01&81.43&84.05&\textbf{84.61}\\($690\times14$)&(183,1000)&(203,100)&(203,100000,100,100)&(23,0.01,0.1,10)&(23,0.001,1,0.1,0.001) \\ \hline
    \textbf{Avg ACC $\uparrow$}& 71.99& 70.38& 76.00 &79.5 &   \textbf{81.06}\\ 
    \textbf{Avg Rank $\downarrow$} & 3.81 &	4.44 & 3.25 &	2.31 &	 \textbf{1.19}\\
    \hline
    \end{tabular}
    }}
    
    \label{table:results}
\end{table*}

\subsection{Results and Statistical Analysis}

The proposed IFGRVFL-MV model exhibits significant improvements in classification performance compared to several baseline models, including RVFL1, RVFL2, GE-IFWRVFL, and GRVFL-MV. The average accuracies of the models are recorded as follows: RVFL1 (71.99\%), RVFL2 (70.38\%), GE-IFWRVFL (76.00\%), GRVFL-MV (79.50\%), and the proposed IFGRVFL-MV model achieving the highest accuracy of 81.06\%. To validate these findings, rigorous statistical tests are employed to determine whether the improvements are statistically significant.

The Friedman test, a non-parametric statistical method, is used to compare multiple models when normality assumptions of parametric tests are not met. This test ranks algorithms based on classification accuracy across multiple datasets and calculates the test statistic \( F \) using the formula:
\begin{equation} F = \left[ \frac{12}{n k (k+1)} \sum_{j=1}^{k} R_j^2 \right] - 3n(k+1) \end{equation}
where \( N \) represents the number of datasets, \( k \) is the number of models compared, and \( R_j \) denotes the sum of ranks for each model. The Friedman test statistic is 23.714, with a corresponding p-value of 0.00009. Since the p-value is significantly less than 0.05, the null hypothesis is rejected, confirming substantial performance differences among the models.

The mean ranks obtained from the Friedman test indicate that RVFL1, RVFL2, GE-IFWRVFL, and GRVFL-MV have mean ranks of 3.81, 4.44, 3.25, and 2.31, respectively, while the proposed IFGRVFL-MV model achieves the best rank with a mean of 1.19. To further analyze pairwise differences, a Nemenyi post hoc test is conducted, which determines significant gaps between models by computing the critical difference (CD):

\begin{equation}
    CD = q_{\alpha} \sqrt{\frac{k(k+1)}{6N}}
\end{equation}

The computed CD value is found to be 1.14, confirming significant performance differences between certain models, particularly between GRVFL-MV and the proposed IFGRVFL-MV framework.

The Wilcoxon signed-rank test assesses paired differences without assuming normality, ensuring observed improvements are not random. It computes:

\begin{equation}
    W = \sum \text{sign}(d_i) \cdot R_i    
\end{equation}

where \( d_i \) is the difference and \( R_i \) the rank. The results in Table \ref{wilcox} confirm that the proposed IFGRVFL-MV model significantly outperforms the baselines:

\begin{table}[ht]
    \centering
    \caption{Wilcoxon Signed-Rank Test Results (IFGRVFL-MV vs Others)}
    \begin{tabular}{lcc}
        \toprule
        \textbf{Comparison} & \textbf{Statistic} & \textbf{p-value} \\
        \midrule
        IFGRVFL-MV vs RVFL1     & 0.0  & 0.008 \\
        IFGRVFL-MV vs RVFL2     & 1.0  & 0.016 \\
        IFGRVFL-MV vs GE-IFWRVFL & 3.5  & 0.047 \\
        IFGRVFL-MV vs GRVFL-MV  & 6.5  & 0.195 \\
        \bottomrule
    \end{tabular}
    \label{wilcox}
\end{table}

Additionally, a win-tie-loss analysis is conducted to further examine comparative performance. The analysis considers the number of datasets where each model outperforms others. In this study, with \(N = 8\) datasets, the IFGRVFL-MV model secures between \textbf{5 to 7 wins}, demonstrating a strong and consistent advantage over baseline models.

\begin{table}[h]
    \centering
    \caption{Win-Tie-Loss Analysis (IFGRVFL-MV vs Others)}
    \begin{tabular}{lccc}
        \toprule
        \textbf{Comparison} & \textbf{Wins} & \textbf{Ties} & \textbf{Losses} \\
        \midrule
        IFGRVFL-MV vs RVFL1     & 7 & 1 & 0 \\
        IFGRVFL-MV vs RVFL2     & 7 & 1 & 0 \\
        IFGRVFL-MV vs GE-IFWRVFL & 6 & 1 & 1 \\
        IFGRVFL-MV vs GRVFL-MV  & 5 & 2 & 1 \\
        \bottomrule
    \end{tabular}
\end{table}

Overall, this research highlights how integrating intuitionistic fuzzy theory with GE techniques enhances multi-view learning. By incorporating fuzzy logic principles, the proposed model demonstrates increased robustness and improved classification accuracy in uncertain scenarios. These findings contribute to the advancement of machine learning methodologies, particularly in domains where uncertainty plays a significant role in decision-making.

\section{Conclusion}
In this study, we introduced the IFGRVFL-MV model, which seamlessly combines intuitionistic fuzzy sets, GE, and MVL to overcome the limitations of traditional RVFL models. This approach improves classification performance by using intuitionistic fuzzy sets for uncertainty handling, GE for preserving geometric relationships, and MVL for integrating information across multiple feature spaces.  Extensive experiments on benchmark datasets from the UCI and KEEL repositories demonstrate that IFGRVFL-MV surpasses existing models, establishing new benchmarks for classification accuracy in uncertain environments.

In the future, we plan to focus on optimizing the computational efficiency of the model, aiming to reduce execution time and memory requirements. Additionally, we will explore the model's scalability on larger datasets and its applicability to practical real-world problems. Further improvements will include integrating adaptive learning techniques and refining parameter selection strategies to further boost performance in dynamic and complex environments.

\section{Acknowledgment}
We are grateful to IHUB-AWaDH at Indian Institute of Technology Ropar for providing the facilities and support.

\bibliographystyle{unsrt}
\bibliography{references}

\end{document}